%% file: main.tex
\documentclass[final]{arxiv}
\usepackage{times}
\usepackage{amsmath}
\usepackage{amssymb}
\usepackage{booktabs}
\usepackage{caption}
\usepackage{colortbl}
\usepackage{cuted}
\usepackage{graphicx}
\usepackage{longtable}
\usepackage{multirow, makecell}
\usepackage{tabularx}
\usepackage[table,xcdraw]{xcolor}
\usepackage[pagebackref=true,breaklinks=true,colorlinks,bookmarks=false]{hyperref}
\usepackage{cleveref}

\input{commands}

\makeatletter
\renewcommand{\paragraph}{%
  \@startsection{paragraph}{4}%
  {\z@}{.5ex \@plus 1ex \@minus .2ex}{-1em}%
  {\normalfont\normalsize\bfseries}%
}
\makeatother

\title{Unsupervised Learning of 3D Object Categories from Videos in the Wild}
\author{Philipp Henzler$^{1}$\footnotemark
\and
Jeremy Reizenstein$^{2}$
\and
Patrick Labatut$^{2}$
\and
Roman Shapovalov$^{2}$ \vspace{.05cm}
\and
\hspace{0.2cm} Tobias Ritschel$^{1}$
\and
Andrea Vedaldi$^{2}$
\and
David Novotny$^{2}$ \hspace{0.2cm} 
\vspace{.2cm}
\and
{\tt\small \{reizenstein,plabatut,romansh,vedaldi,dnovotny\}@fb.com} \and \hspace{3.cm} {\tt\small \{p.henzler,t.ritschel\}@cs.ucl.ac.uk} \hspace{3.cm}
\vspace{.3cm}
\and 
\hspace{0.4cm} $^{1}$University College London \and
\vspace{.1cm}
$^{2}$Facebook AI Research
}

\begin{document}
\maketitle
\input{fig-splash}

\footnotetext[1]{Work completed during an internship at Facebook AI Research.}

\begin{abstract}\vspace{-1.0em}
Our goal is to learn a deep network that, given a small number of images of an object of a given category, reconstructs it in 3D.
While several recent works have obtained analogous results using synthetic data or assuming the availability of 2D primitives such as keypoints, we are interested in working with challenging real data and with no manual annotations.
We thus focus on learning a model from multiple views of a large collection of object instances.
We contribute with a new large dataset of object centric videos suitable for training and benchmarking this class of models.
We show that existing techniques leveraging meshes, voxels, or implicit surfaces, which work well for reconstructing isolated objects, fail on this challenging data.
Finally, we propose a new neural network design, called \emph{warp-conditioned ray embedding} (WCR), which significantly improves reconstruction while obtaining a detailed implicit representation of the object surface and texture, also compensating for the noise in the initial SfM reconstruction that bootstrapped the learning process.
Our evaluation demonstrates performance improvements over several deep monocular reconstruction baselines on existing benchmarks and on our novel dataset.
For additional material please visit:
{\small \url{https://henzler.github.io/publication/unsupervised_videos/}}.

\end{abstract}
\input{intro}
\input{related}
\input{method}
\input{experiments}

\input{conclusions}

{\small%
\bibliographystyle{ieee_fullname}%
\bibliography{refs}
}

\clearpage
\appendix
\input{supplemental}

\end{document}

%% file: commands.tex
\usepackage{soul}
\usepackage{microtype}
\usepackage{wrapfig}
\usepackage{fancyhdr}

\def\figurePath{figures/}
\def\myfigure#1#2{\begin{figure}[htb]\centering\includegraphics*[width = \linewidth]{\figurePath#1}\centering\caption{#2}\label{fig:#1}\end{figure}}

\def\mycfigure#1#2{\begin{figure*}[t]\centering\includegraphics*[clip, width = \linewidth]{\figurePath#1}\centering\caption{#2}\label{fig:#1}\end{figure*}}

\def\myhfigure#1#2{\begin{figure*}[h!]\centering\includegraphics*[clip, width = \linewidth]{\figurePath#1}\centering\caption{#2}\label{fig:#1}\end{figure*}}

\renewcommand{\eg}{e.\,g., }
\renewcommand{\ie}{i.\,e., }
\renewcommand{\etal}{et~al.\ }

\newcommand{\refFig}[1]{Fig.~\ref{fig:#1}}

\usepackage{pifont}

\usepackage{adjustbox}
\newcolumntype{R}[2]{%
    >{\adjustbox{angle=#1,lap=\width-(#2)}\bgroup}%
    l%
    <{\egroup}%
}

\soulregister\ref7
\soulregister\cite7
\soulregister\refFig7
\soulregister\cite7
\soulregister\ref7
\soulregister\pageref7
\soulregister\shortcite7
\soulregister\eg0
\soulregister\ie0
\soulregister\etal0

\DeclareGraphicsExtensions{.png,.jpg,.pdf,.ai,.psd}
\newcommand{\mysection}[2]{\section{#1}\label{sec:#2}}
\newcommand{\mysubsection}[2]{\subsection{#1}\label{sec:#2}}

\newcommand{\bc}{\mathbf{c}}
\newcommand{\x}{\mathbf{x}}

\newcommand{\z}{\mathbf{z}}

\newcommand{\br}{\mathbf{r}}

%% file: fig-splash.tex
\begin{strip}
\vspace{-3.0em}
\begin{center}
\includegraphics[width=\linewidth]{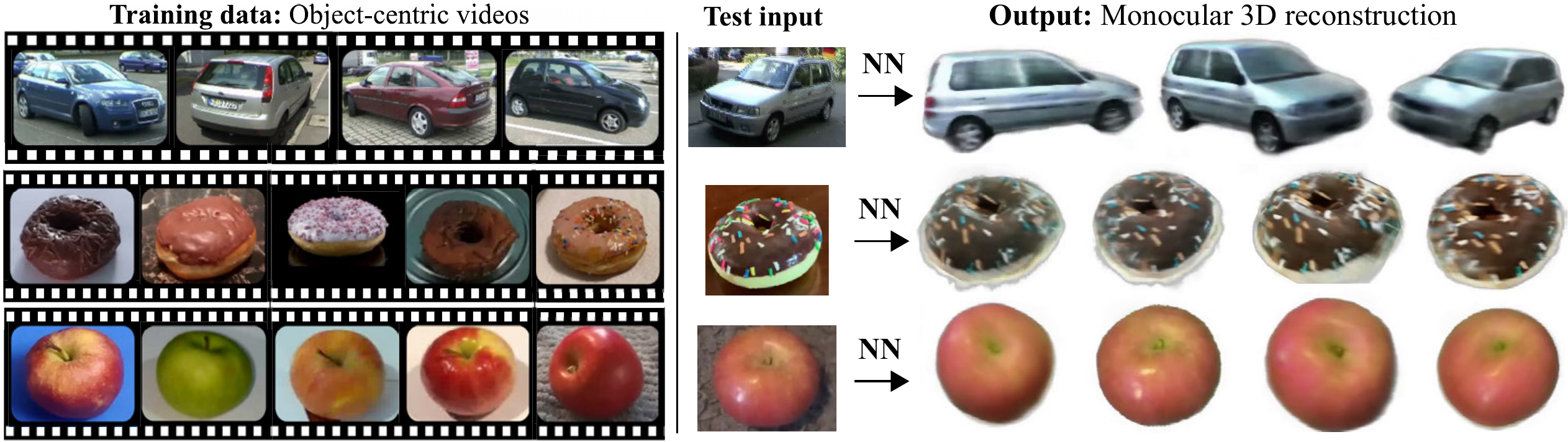}
\captionof{figure}{%
We present a novel deep architecture that contributes \textit{Warp-conditioned Ray Embedding (WCR)} to reconstruct and render new views (right) of object categories from one or few input images (middle).
Our model is learned automatically from videos of the objects (left) and works on difficult real data where competitor architectures fail to produce good results.%
}\label{f:splash}%
\end{center}
\end{strip}

%% file: intro.tex
\mysection{Introduction}{Introduction}

Understanding and reconstructing categories of 3D objects from 2D images remains an important open challenge in computer vision.
Recently, there has been progress in using deep learning methods to do so but, due to the difficulty of the task, these methods still have significant limitations.
In particular, early efforts focused on clean synthetic data such as ShapeNet~\cite{chang2015shapenet}, further simplifying the problem by assuming the availability of several images of each object instance, knowledge of the object masks, object-centric viewpoints, etc.
Methods such as~\cite{fan2017point,choy20163d,rezende2016unsupervised,tulsiani2017multi, kar2017learning} have demonstrated that, under these restrictive assumptions, it is possible to obtain high-quality reconstructions, motivating researchers to look beyond synthetic data.

Other methods have attempted to learn the 3D shape of object categories given a number of independent views of real-world objects, such as a collection of images of different birds.
However, in order to simplify the task, most of them use some form of manual or automatic annotations of the 2D images.
We seek to relax these assumptions, avoiding the use of manual 2D annotations or \emph{a priori} constraints on the reconstructed shapes. 

When it comes to high-quality general-purpose reconstructions, methods such as~\cite{mildenhall2020nerf,liu2020neural,martinbrualla2020nerfw,niemeyer2020differentiable,yariv2020multiview} have demonstrated that these can be obtained by training a deep neural network given only multiple views of a scene or object without manual annotations or particular assumptions on the 3D shape of the scene.
Yet, these techniques can only learn a single object or scene at a time, whereas we are interested in modelling entire categories of 3D objects with related but different shapes, textures and reflectances.
Nevertheless, the success of these methods motivates the use of multi-view supervision for learning collections of 3D objects.

In this paper, our first goal is thus to learn 3D object categories given as input multiple views of a large collection of different object instances.
To the best of our knowledge, this is the first paper to conduct such a large-scale study of reconstruction approaches applied to learning 3D object categories from real-world 2D image data.
Unfortunately, existing datasets for 3D category understanding are either small or synthetic.
Thus, our first contribution is to introduce a new dataset of videos collected `in the wild' by Mechanical Turkers (\cref{fig:dataset}).
These videos capture a large number of object instances from the viewpoint of a moving camera, with an effect similar to a turntable.
Viewpoint changes are estimated with high accuracy using off-the-shelf Structure from Motion (SfM) techniques.
We collect hundreds of videos of several different categories.

Our second contribution is to assess current reconstruction technology on our new `in the wild' data.
For example, since each video provides several views of a single object with known camera parameters, it is suitable for an application of recent methods such as NeRF \cite{mildenhall2020nerf}, and we find that learning \emph{individual videos} works very well, as expected.
However, we show that a direct application of such models to several videos of different but related objects is \emph{much} harder.
In fact, we experiment with related representations such as voxels and meshes, and find that they also do not work well if applied na\"{\i}vely to this task.
This is true even though reconstructions are focused on a single object at a time --- thus disregarding the background --- suggesting that these architectures have a difficult time at handling even relatively mild geometric variability.

Our final contribution is to propose a novel deep neural network architecture to better learn 3D object categories in such difficult conditions.
We hypothesize that the main challenge in extending high-quality reconstruction techniques, that work well for single objects, to object categories is the difficulty of absorbing the geometric variability that comes in tackling many different objects together.
An obvious but important source of variability is \emph{viewpoint}:
given only real images of different objects, it is not obvious how these should align in 3D space, and a lack of alignment adds to the variability that the model must cope with.
We address this issue with a novel idea of \textit{Warp-Conditioned Ray Embeddings (WCR)}, a new neural rendering approach that is far less sensitive to inaccurate 3D alignment in the input data.
Our method modifies previous differentiable ray marchers to pool information at variable locations in input views, conditioned on the 3D location of reconstructed points.

With this, we are able to train deep neural networks that, given as input a small number of images of new object instances in a given target category, can reconstruct them in 3D, including generating high-quality new views of the objects.
Compared to existing state-of-the-art reconstruction techniques, our method achieves better reconstruction quality in challenging datasets of real-world objects.

\input{fig-overview}

%% file: fig-overview.tex
\begin{figure*}[ht]%
\centering%
\includegraphics[width=\textwidth]{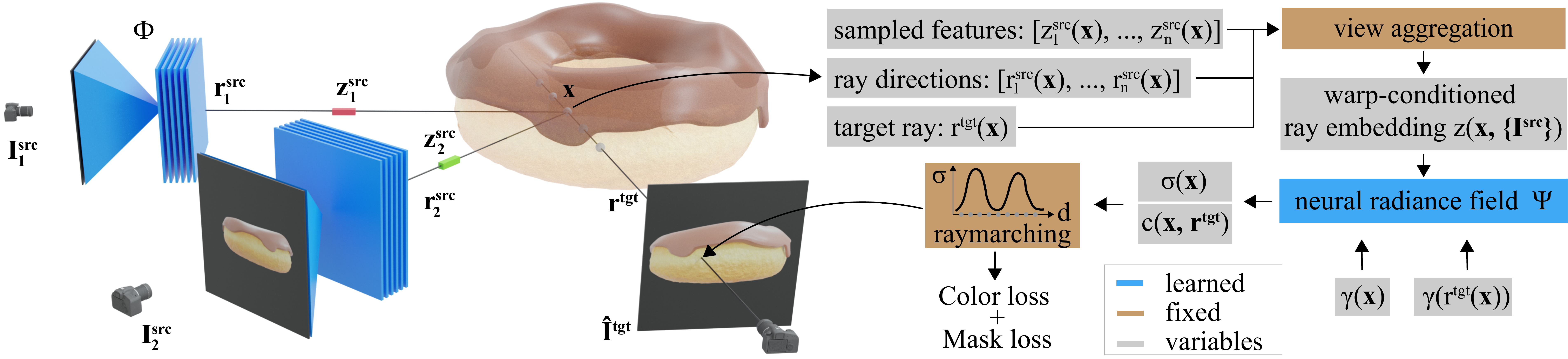}%
\vspace{-0.7em}%
\caption{Our method takes as input an image and produces per pixel features using a U-Net $\Phi$. We then shoot rays from a target view and retrieve per-pixel features from one or multiple source images. Once all spatial feature vectors are aggregated into a single feature vector (see \Cref{s:image_encoding} for more details), we combine them with their harmonic embeddings and pass them to an MLP yielding per location colors and opacities. Finally, we use differentiable raymarching to produce a rendered image.\label{Overview}}%
\vspace{-1.0em}%
\end{figure*}%

%% file: related.tex
\mysection{Related Work}{RelatedWork}

Our work is related to many prior papers that leveraged deep learning for 3D reconstruction.

\paragraph{Learning synthetic 3D object categories.}

Early deep learning methods for 3D reconstruction focused on clean synthetic datasets such as ShapeNet~\cite{chang2015shapenet}. 
Fully supervised methods \cite{choy20163d,girdhar2016learning} 
mapped 2D images to 3D voxel grids.
Follow-up methods proposed several alternatives:
\cite{fan2017point, yang2019pointflow} predict a point clouds, Park et al. \cite{park2019deepsdf,atzmon2020sal} label each 3D point with its signed distance to the nearest surface point, \cite{mescheder2019occupancy,chen2019learning} predict binary per-point occupancies, \cite{genova2019learning,genova2019deep} proposed more structured occupancy functions, and
\cite{gkioxari19mesh,wang2018pixel2mesh} reconstruct meshes from single views.
All aforementioned methods require full supervision in form of images and corresponding 3D CAD models. 
In contrast, our method requires only a set of videos of an object category captured from a moving camera.

Methods that avoid 3D supervision project 3D shapes to 2D images using differentiable rendering, allowing for image space optimization instead of 3D~
\cite{rezende2016unsupervised, yan2016perspective, tulsiani2017multi,kar2017learning, insafutdinov2018unsupervised,tulsiani2018multi}. 

\paragraph{Learning 3D object categories in the wild.}

Early reconstruction methods for 3D object categories used Non-Rigid SfM (NR-SfM) applied to 2D keypoint annotations~\cite{cashman2013shape,vicente2014reconstructing,carreira2015virtual}.
CMR~\cite{kanazawa18learning} used NR-SfM and 2D keypoints to initialize the camera poses on the CUB \cite{WahCUB_200_2011} dataset based on differentiable mesh rendering \cite{kato2018neural,liu19soft,chen2019learning}.
The texturing model of CMR was improved in DIB-R \cite{chen2019learning}.

Instead of assuming knowledge of pose, \cite{kulkarni19canonical,kulkarni20articulation-aware,goel2020shape} assume a deformable 3D template.
PlatonicGAN~\cite{henzler19escaping} enables template-free 3D reconstruction via differentiable emission-absorption raymarching, but requires knowledge of the camera-pose distribution.

Similarly, \cite{wu2020unsupervised} does not require pose supervision, but it has been demonstrated only for limited viewpoint variations.
Li et al.~\cite{li2020self} do not assume camera poses as input, but use the self-supervised semantic features of \cite{hung2019scops} as a proxy for 2D keypoints as well as further constraints such as symmetry to help the reconstruction.
We avoid such constraints for the sake of generality.
Exploiting the StyleGAN~\cite{karras19stylegan} latent space, Zhang et al.~\cite{zhang2020image} only require very few manual pose annotations.
Our method, furthermore, does not require keypoint or pose supervision; instead, it recovers scene-specific camera poses automatically by analyzing camera motion.
\cite{novotny17learning,novotny2018capturing} canonically align point clouds by only supervising with relative pose, but only learn a shape model.

Generative models trained in the wild were proposed in \cite{gadelha20173d,nguyen-phuoc19hologan}. 
While these methods can `hallucinate' high-quality images, they, unlike us, are unable to also perform reconstruction of the objects given an image as input.

\paragraph{Implicit representation of 3D scenes.}

NeRF~\cite{mildenhall2020nerf} has raised the interest in neural scene representation due to its high-quality output, inspired by positional encoding proposed in \cite{vaswani2017attention} and differentiable volume rendering from \cite{henzler19escaping,tulsiani2017multi}.
NSVF~\cite{liu2020neural} combined NeRF and voxel grids to improve the scalability and expressivity of the model whereas Yariv et al~\cite{yariv2020multiview} uses sphere tracing to render signed distance fields.
GRAF~\cite{schwarz2020graf} extended NeRF to allow learning category-specific image generators, but do not perform reconstruction, which is our goal.
Our method is inspired by NeRF, however, we learn a model of a whole object category, rather than a single scene or object.

Recent works, \cite{huang2018deep,saito19pifu,saito2020pifuhd,yu2020pixelnerf,wang2021ibrnet} utilize sampled per-pixel encodings similar to us. 
\cite{yu2020pixelnerf} averages features over multiple views and \cite{wang2021ibrnet} learns to interpolate between views in an IBR fashion \cite{hedman18deepblending,buehler2001unstructured} which prevents inpainting unseen areas.
Our method aggregates latent encodings, which allows for representing unseen areas.
Furthermore, we observed that simply averaging features from significantly different viewpoints, as done in \cite{yu2020pixelnerf}, hurts performance.
We thus propose to aggregate depending on view angles.

%% file: method.tex
\mysection{Method}{Method}

\paragraph{Overview.}

The goal of our method is to learn a model of a 3D object category from a dataset $\{\mathcal{V}^p\}_{p=1}^{N_\text{video}}$ of video sequences.
Each video $\mathcal{V}^p = (I_t^p)_{0\leq t < T^p}$ consists of $T^p \in \mathbb{N}$ color frames $I_t^p \in \mathbb{R}^{3 \times H \times W}$.
While we do not use any manual annotations for the videos, we do pre-process them using a Structure-from-Motion algorithm (COLMAP~\cite{schoenberger2016sfm}).
In this manner, for each video frame $I_t^p$, we obtain sequence-specific camera poses $g_t^p \in SE(3)$ and the camera instrinsics $K_t^p \in \mathbb{R}^{3 \times 3}$.
We further obtain a segmentation mask $m_t^p \in R^{1 \times H \times W}$ of the given category using Mask-RCNN \cite{he17mask}.

The model parametrizes the appearance and geometry of the object in each video with an implicit surface map $\Psi$:
$$
\Psi: \mathbb{R}^3 \times \mathbb{S}^2 \times \mathcal{Z} \rightarrow \mathbb{R}^3 \times \mathbb{R}_+ ~\quad
\Psi(\x, \br, \z) = (\bc, \sigma),
$$
which labels each 3D scene point $\x \in \mathbb{R}^3$ and viewing direction $\br \in \mathbb{S}^2$ with an RGB triplet $\bc(\x, \br, \z)\in\mathbb{R}^3$ and an occupancy value $\sigma(\x, \z)\in(0, 1]$ representing the opaqueness of the 3D space.
Furthermore, the implicit function $\Psi$ is conditioned on a latent code $\z \in \mathcal{Z}$ that captures the factors of variation of the object.
By changing $\z$ we can adjust the occupancy field to represent shapes of different objects of a visual category.
As described in~\cref{s:image_encoding}, the design of the latent space $\mathcal{Z}$ is crucial for the success of the method.

While we use video sequences to train the model, at test time we would like to reconstruct any new object instance from a small number of images.
To this end, we learn an encoder function
$$
\Phi : \mathbb{R}^{3\times H\times W\times N_\text{src}} \rightarrow \mathcal{Z},
$$
that takes a number of input \emph{source} images
$
\{ I^\text{src}_1, \dots, I^\text{src}_{N_\text{src}} \}
$
of the new instance and produces the latent code $\z \in \mathcal{Z}$.

Given a known target view (different view than the source images) we render the implicit surface to form a color image $\hat I^\text{tgt} \in \mathbb{R}^{3\times H\times W}$ and minimize the discrepancy between the rendered $\hat I^\text{tgt}$ and the masked ground truth image $I^\text{tgt}$.

In the following, we describe the main building blocks of our method.
The rendering step follows Emission-Absorption raymarching~\cite{max95optical,henzler19escaping,mildenhall2020nerf,tulsiani2018multi} as detailed in \cref{s:diff_render}. \Cref{s:neural_surface} describes the specifics of the surface function $\Psi$, and \cref{s:image_encoding} introduces the main technical contribution --- a novel Warp-Conditioned Ray Embedding that defines the image encoder $\Phi$.

\subsection{Implicit surface rendering}\label{s:diff_render}

In order to render a target image $\hat I^\text{tgt}$, we emmit a ray from the camera center through each pixel, assigning the color of ray's first `intersection' with the surface to the respective pixel.
Formally, let $\Omega =\{0,\dots, W-1\} \times \{0,\dots,H-1\}$ be an image grid, $u \in \Omega$ the index of a pixel, and $Z \in \mathbb{R}_+$ a depth value.
Following the ray from the camera center through $u$ to depth $Z \geq 0$ results in the 3D point:
$
\bar\x(u,Z)
=
Z\cdot K^{-1} [ u^\top~1]^\top
$,
where $K\in\mathbb{R}^{3\times 3}$ are the camera intrinsics. %
The camera's pose is given by an Euclidean transformation $g^\text{tgt}\in SE(3)$, where we use the convention that 
$\bar\x = g^\text{tgt}(\x)$
maps points $\x$ expressed in the world reference frame to points $\bar\x$ in camera coordinates.

In order to determine the color of a pixel $u\in\Omega$, we then `shoot' a ray seeking the surface intersection.
To do so, we sample points 
$\mathcal{X}_u = \left(\x(u, Z_i)\right)_{i=0}^{N_Z+1}$ 
for depth values $Z_0\leq \cdots \leq Z_{N_Z}$ obtaining their colors and occupancies:
\begin{equation}\label{e:ray-samples}
  (\bc_i, \sigma_i) = \Psi(\x(u, Z_i),\br, \z),~~~i=0,\dots,N_Z.
\end{equation}
The probability of the ray \emph{not} intersecting the surface in the interval $(Z_{i+1},Z_i]$ is set to 
$T_i = e^{-(Z_{i+1}-Z_i)\sigma_i(\x(u, Z_i), \z)}$
(transmission probability).
Summing over all possible intersections $Z_0, \dots, Z_i$, the probability $p(Z=Z_i|u)$ of a ray terminating at depth $Z_i$ is thus defined as:
$$
p(Z=Z_i|u)
=
\left(\prod_{j=0}^{i-1} T_j\right)
\left(1 - T_i\right),
~~
\hat m_u
=
1 - \prod_{i=0}^{N_Z-1}
T_i,
$$
with the overall probability of intersection $\hat m_u$.
Given the distributions of ray-termination probabilities $p(Z|u)$, the rendered color $\hat \bc_u(\mathcal{X}_u, \br, \z) \in \mathbb{R}^3$ 
and opacity 
$\hat \sigma_u(\mathcal{X}_u, \z) \in \mathbb{R}$ 
are defined as an expectation over the outputs of the implicit function within the range $[0, \dots, N_Z-1]$:
$$
\hat \bc_u
=
\sum_{i=0}^{N_Z-1} p(Z=Z_i|u) \bc_i,
~~
\hat \sigma_u
=
\sum_{i=0}^{N_Z-1} p(Z=Z_i|u) \sigma_i.
$$
Since we are only interested in rendering the interior of the object, the colors $\bc_u$ are softly-masked with $\hat m_u$ leading to the final target image render $\hat I^\text{tgt} \in \mathbb{R}^{3\times H\times W}$:
\begin{equation}\label{e:tgt}
  \hat I^\text{tgt} = I(g^\text{tgt}, \z)= \hat m \odot \hat \bc.
\end{equation}
Note that the reconstruction depends on the target viewpoint $g^\text{tgt}$ and the object code $\z$, which is viewpoint independent.

\subsection{Neural implicit surface}\label{s:neural_surface}

Next, we detail the implicit surface function $\Psi$.
Similar to previous methods~\cite{mildenhall2020nerf,niemeyer19occupancy,mescheder2019occupancy}, we exploit the representational power of deep neural networks and define $\Psi$ as a deep multi-layer perceptron (MLP):
$
(\bc, \sigma) = \Psi_\text{nr}(\x, \br, \z).
$
The network $\Psi_{\text{nr}}$ follows a design similar to~\cite{mildenhall2020nerf}.
In particular, the world-coordinates $\x$ are preprocessed with the \emph{harmonic encoding} $\gamma_{N_{f}^x}(\x) = [\sin(\x), \cos(\x), \dots, \sin(2^{N_{f}^x} \x), \cos(2^{N_{f}^x} \x)] \in \mathbb{R}^{2 N_{f}^x} $ before being input to the first layer of the MLP\@.
In order to enable modelling of viewpoint dependent color variations, we further use the harmonic encoding of the target ray direction $\gamma_{N_f^r}(\br^\text{tgt}(\x)) \in \mathbb{R}^{2 N_f^r}$ as input (see \Cref{Overview}).

\subsection{Warp-conditioned ray embedding}\label{s:image_encoding}

An important component of our method is the design of the latent code $\z$.
A na\"{\i}ve solution is to first map a source image $I^\text{src}$ to a $D$-dimensional vector $\z_\text{CNN} = \Phi_\text{CNN}(I^\text{src}) \in \mathbb{R}^D$ with a deep convolutional neural network $\Phi_\text{CNN}$, followed by appending a copy of $\z_\text{CNN}$ to each positional embedding $\gamma(\x)$ to form an input to the neural occupancy function $\Psi_{\text{nr}}$.
This approach, successfully utilized in~\cite{tulsiani2018multi,liu19soft} for synthetic datasets where the training shapes are approximately rigidly aligned, is however insufficient when facing more challenging in-the-wild scenarios.

To show why there is an issue here, recall that our inputs are \emph{videos} $\mathcal{V}^p$ of different object instances, each consisting of a sequence $(I_t^p)_{0\leq t < T^p}$ of video frames, together with 
viewpoint transformations $g^p_t \in SE(3)$ recovered by SfM.
Crucially, due to the global coordinate frame and scaling ambiguity of the SfM reconstructions~\cite{hartley2003multiple}, there is no relationship between the camera positions $g^p$ and $g^q$ reconstructed for two different videos $p\not= q$.
Even two identical videos $\mathcal{V}^p = \mathcal{V}^q$, reconstructed using SfM from two different random initializations, will result in two different sets of cameras
$ 
(g_t^p)_{0\leq t < T^p}
$,
$
(g_t^q = g^\star g_t^p)_{0\leq t < T^p}
$,
related by an unknown similarity transformation $g^\star \in S(3)$.
Since the frames $I_t^p= I_t^q$ are identical, the reconstruction network $\Phi_\text{CNN}$ must assign to them identical codes:
$
\z_{\text{CNN},t} = \z_\text{CNN,t}^p
= 
\Phi_\text{CNN}(I_t^p)
=
\Phi_\text{CNN}(I_t^q)
=
\z_\text{CNN,t}^q
$.
Plugging this in~\cref{e:tgt}, means that two identical frames are reconstructed from the same code $\z_{\text{CNN},t}$ but two different viewpoints $g^p_t\not=g^q_t$:
$
\hat I^p_t
=
I(g^p_t, \z_{\text{CNN},t})
=
I(g^q_t, \z_{\text{CNN},t})
=
\hat I^q_t
$.
While of course we do not work with identical copies of the same videos, this extreme case demonstrates a fundamental issue with the na\"{\i}ve model, where different object instances must be reconstructed with respect to unrelated viewpoints.

We can partially tackle this issue by using a variant of~\cite{novotny16learning} to approximately align the viewpoint of different video sequences before training (see supplemental).

Next, we introduce a more fundamental change to the model that also helps addressing this issue.
The idea is to change the implicit surface~\eqref{e:ray-samples}
\begin{equation}\label{e:pre-wcr}
  \Psi_\text{WCR}(\x, \z(\x)),
\end{equation}
such that the code $\z$ is a \emph{function of the queried ray point $\x$} in world coordinates. Given a source image $I^\text{src}_t$ with viewpoint $g_t$, the projection of this point in the image is:
$
u_t(\x) = \pi_t(\x) = \pi(K g_t \x)
$
where $\pi$ denotes the perspective projection operator $\mathbb{R}^3 \rightarrow \Omega$.
In particular, if $\x$ is also a point on the surface of the object, then $u_t(\x)$ is the image of the corresponding point in the source view $I^\text{src}_t$.

More specifically, we task a convolutional neural network $\Phi$ to map the image $I^\text{src}_t$ to a feature field $\Phi(I^\text{src}_t) \in \mathbb{R}^{D\times H\times W}$ (see supplementary for details).
In this way, for each pixel $u_t$ in the source view, we obtain a corresponding embedding vector $\Phi(I_t)[u_t(\x)]$ (using differentiable bilinear interpolation $[\cdot]$):
\begin{equation}\label{e:wcr}
  \z_t(\x) = \Phi(I_t)[\pi_t(\x)] \in \mathbb{R}^D,
\end{equation}
and call it \textbf{Warp-Conditioned Ray Embedding (WCR)}.

Intuitively, as shown in~\cref{Overview}, by using~\cref{e:pre-wcr,e:wcr} during ray marching, the implicit surface network $\Psi_\text{WCR}$ can pool information from relevant 2D locations $u_t$ in the source view $I_t^\text{src}$.
Importantly, this occurs in a manner which is invariant to the global viewpoint ambiguity.
In fact, if the geometry is now changed by the application of an arbitrary similarity transformation $g^\star$, then the 3D point changes as $\x' = g^\star \x$, but the viewpoint also changes as $g_t' = g_t (g^\star)^{-1}$, so that $g_t' \x' = g_t' (g^\star)^{-1} g^\star \x = g_t \x$ and the encoding of the points $\x$ and $\x'$ is the same:
$
\Phi(I_t)[\pi_t(\x)] = \Phi(I_t)[\pi_t'(\x')]
$
Finally, note that the network~\cref{e:pre-wcr} combines two sources of information:
(1) codes $\z(\x)$ that capture the appearance of each point in a manner which is invariant from the global coordinate transforms; and
(2) the absolute location of the 3D point $\x$ (internally encoded by using position-sensitive coding $\gamma(\x)$).
The combination of 1) and 2) above allows to resolve misalignments by localizing the implicit surface equivariantly with changes of the global coordinates.

\paragraph{Multi-view aggregation.}%
Having described WCR for a single source image we now extend to the more common case with multiple source images.
For a set of source views 
$\{I_t^\text{src}\}_{t=1}^{N_\text{src}}$ 
with their warp-conditioned embeddings $\z^\text{src}_t(\x)$, source rays $\br^\text{src}_t(\x)$, and the target ray $\br^\text{tgt}(\x)$ (see \Cref{Overview}), we calculate the aggregate WCR 
$\z(\x, \{I_t^\text{src}\})$:

\begin{align*}\label{e:wcr_aggregate}
\z(\x, \{I_t^\text{src}\})& = 
\text{cat}\big(\\&
\z^\mu(\x, \{I_t^\text{src}\}), 
\z^\sigma(\x, \{I_t^\text{src}\}), 
\z_\text{CNN}(\{I_t^\text{src}\})
\big),
\end{align*}

as a concatenation ($\text{cat}$) of the angle-weighted mean and variance embedding $\z^\mu \in \mathbb{R}^D$ and $\z^\sigma \in \mathbb{R}_+$ respectively, and a plain average  $\z_\text{CNN} = N_\text{src}^{-1} \sum_t \z_{\text{CNN}, t}$ over global source embeddings $\z_{\text{CNN}, t}$.

The mean 
$\z^\mu(\x,  \{I_t^\text{src}\})=\sum_{t=1}^{N_\text{src}} w_t(\x) \z^{src}_t(\x)$
is a weighted average of the source embeddings $\z^{src}_t(\x)$
with the weight $w_t(\x)$ defined as 
$$
w_t(\x) = {W(\x)}^{-1} (1 + \br^\text{src}_t(\x) \cdot \br^\text{tgt}(\x)).
$$
$W(\x) = \sum_{t=1}^{N_\text{src}} w_t(\x)$ is a normalization constant ensuring the weights integrate to 1. 
This gives more weight to the source-view features that are imaged from a viewpoint which is closer to the target view. 
The variance embedding $\z^\sigma\in\mathbb{R}_+$ is defined analogously as an average over dimension-specific $w_t(\x)$-weighted standard deviations of the source embedding set
$\{\z^\text{src}_t(\x)\}_{t=1}^{N_\text{src}}$.

\subsection{Overall learning objective}\label{s:optimization}

For training, we optimize the loss
$
\mathcal L = \lambda \mathcal L_{\text{mask}} + \mathcal L_{\text{rgb}} $
where $\lambda = 0.05$.
$\mathcal L_{\text{mask}}$ is defined as the binary cross-entropy between the rendered opacity and ground truth mask. For the appearance loss $\mathcal L_{\text{rgb}}$ we use the mean-squared error between the masked target view and our rendering.

%% file: experiments.tex
\begin{figure}[h]
\begin{center}
\includegraphics[width=\linewidth]{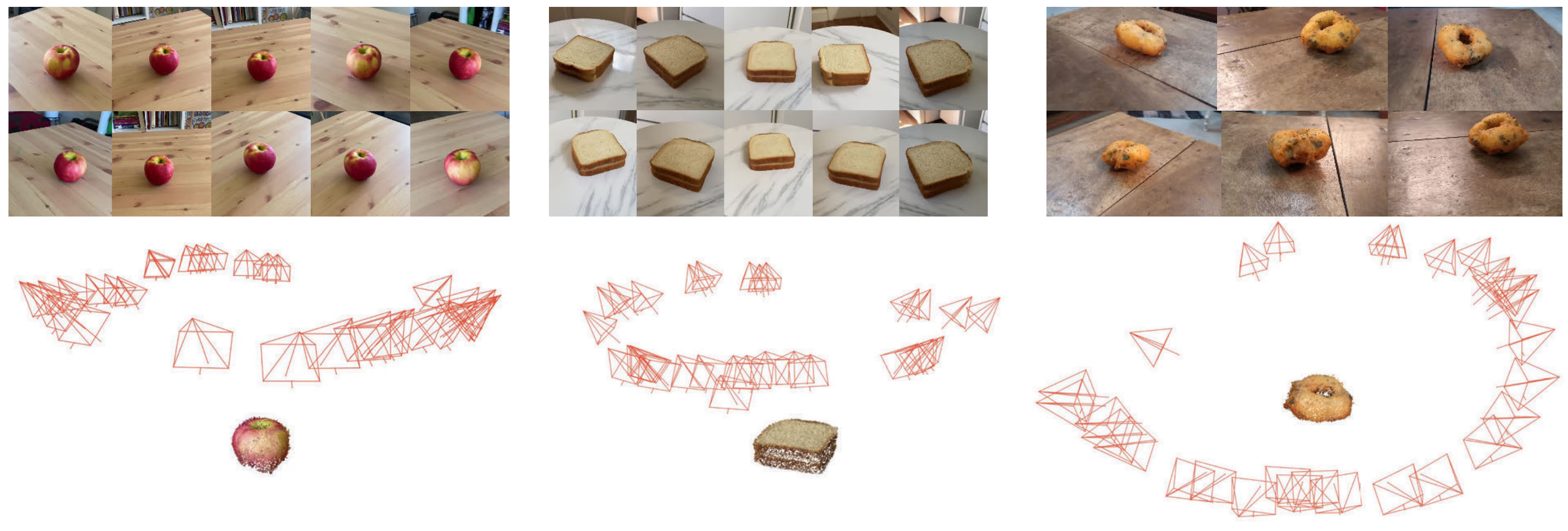}%
\vspace{-0.7em}%
\caption{%
In order to study learning 3D object categories in the wild, we crowd-sourced a large collection of object-centric videos from Amazon Mechanical Turk.
The top row shows frames from three example videos, the bottom two rows show SfM reconstructions of the videos together with tracked cameras.\label{fig:dataset}}%
\end{center}%
\vspace{-1.2em}%
\end{figure}

\mysection{Experiments}{Experiments}

We discuss implementation details, data and evaluation protocols (\cref{s:data}) and assess our method and baselines on the tasks of novel-view synthesis and depth prediction.

\input{table-results}

\paragraph{Implementation details.}

As noted in~\cref{s:image_encoding}, although WCR is in principle capable of dealing with the scene misalignments by itself, we found it beneficial to approximately ``synchronize'' the viewpoints of different videos in pre-processing, using a modified version of the method from~\cite{novotny16learning}. 
First, we use the scene point clouds from SfM to register translation and scale by centering (subtracting the mean) and dividing by average per-dimension variance, resulting in adjusted viewpoints $\bar g_t$.
We then proceed with training the rotation part of the viewpoint factorization branch of the VpDR network from~\cite{novotny16learning}, in order to align the rotational components of the viewpoints.

\mysubsection{AMT Objects and other benchmarks}{Datasets}\label{s:data}

One of our main contributions is to introduce the \textbf{AMT Objects} dataset, a large collection of object-centric videos that we collected (\cref{fig:dataset}) using Amazon Mechanical Turk.
The dataset contains 7 object categories from the MS COCO classes~\cite{lin14microsoft}: apple, sandwich, orange, donut, banana, carrot and hydrant. 
For each class, we ask Turkers to collect a video by looking `around' a class instance, resulting in a turntable video.
For reconstruction, we uniformly sampled 100 frames from each video, discarding any video where COLMAP pre-processing was unsuccessful.
The dataset contains 169-457 videos per class.
For each class, we randomly split videos into training and testing videos in an 8:1 ratio.

We also consider the \textbf{Freiburg Cars}~\cite{nima2015unsupervised}, consisting of 45 training and 5 testing videos of various parked cars.

\myhfigure{SingleViewResults}{\textbf{Monocular reconstruction on Freiburg Cars and AMT Objects.}
In reach row, a single source image (1st column) is processed by one of the evaluated methods (Mesh, Voxel, MLP+Voxel, MLP, \textbf{Ours} - columns 2 to 6) to generate a prescribed target view (last column).
We show results on the test split.%
} 

For every video, we define three disjoint sets of frames on which we either train or evaluate: 
(1) \emph{train-train},
(2) \emph{train-test} and
(3) \emph{test}.
For each training video, we form the train-test set by randomly selecting 16 frames and a disjoint train-train set containing the complement of train-test. 
While the train-train frames are utilized for training, the train-test frames are never seen during training and only serve for evaluation.
The evaluation on the test set is the most challenging since it is conducted with views of previously unseen object instances.

\paragraph{Evaluation protocol.}

Recall that, at test time, our network takes as input a certain number of source images $I^\text{src}$ and reconstructs a target image $\hat{I}^\text{tgt}$ seen from a different viewpoint.
We assess the view synthesis and depth reconstruction quality of this prediction.
To this end, for each object category, we randomly extract a batch of 8 different images from the \emph{train-test} and \emph{test} respectively. To increase view variability we repeat this process 5 times for every object.
For each batch one of the images is picked as a target image $I^\text{tgt}$ and from the remaining images we individually select 1,3,5,7 images and perform the forward pass to generate $\hat{I}^\text{tgt}$ for each selection.

In order to assess the quality of view synthesis, we calculate the \textbf{$\ell_1^{\text{RGB}}$} error, between the target and predicted image. We also use the \textbf{$\ell_1^{\text{VGG}}$} perceptual metric, which computes the $\ell_1$ distance between the two images encoded by means of the VGG-19 network~\cite{simonyan15very} pretrained on ImageNet.
For depth reconstruction, we compute the \textbf{$\ell_1^{\text{Depth}}$} distance between ground truth depth map (obtained from COLMAP SfM) and the predicted one in the target view.
Finally we report Intersection-over-Union (\textbf{IoU}) between the predicted object mask and the object mask obtained by Mask-RCNN in the target view.

\input{table-viewablation}

\myhfigure{MultiViewResult}{%
\textbf{Reconstruction with multiple source views.}
For each object, the top row shows all available source images (columns 1-7) for a given target image (top right).
The bottom row contains results conditioned on 1, 3, 5 or 7 source images. 
In addition to the rendered new RGB views we also provide shaded surface renderings.%
}

\mysubsection{Baselines}{baselines}

In this section we detail the baselines we compare with.
The first is \textbf{MLP}, corresponding to a na\"{\i}ve version of the latent global encoding $\z_\text{CNN}$ already discussed in \cref{s:image_encoding}.
Here, the $N^\text{src}$ source images $\{I^{src}_t\}_{t=1}^{N^\text{src}}$ are first independently mapped to embedding vectors $\{\z_t \in \mathbb{R}^{256}\}_{t=1}^{N^\text{src}}$ by a ResNet50~\cite{he2016deep} encoder and subsequently averaged to form an encoding of the object $\z_\text{CNN} = \frac{1}{N^\text{src}} \sum_{t=1}^{N^\text{src}} \z_t$.
A copy of $\z_\text{CNN}$ is then concatenated to each positional embedding $\gamma(\x)$ of each target ray point $\x$. 
MLP renders with the EA ray marcher (\cref{s:diff_render}).

The second baseline is \textbf{Voxel}, which closely resembles~\cite{tulsiani2017multi}.
This uses the same encoding scheme as \textbf{MLP}, but differs by the fact that the object is represented by a voxel grid.
Specifically, $\z_\text{CNN}$ is decoded with a series of 3D convolution-transpose layers to a $128^3$ voxel grid containing RGB and opacity values.
\textbf{Voxel} also renders with EA.

Next, \textbf{Voxel+MLP} is inspired by Neural Sparse Voxel fields~\cite{liu2020neural} and marries NeRF~\cite{mildenhall2020nerf} with voxel grids.
As in \textbf{Voxel}, $\z_\text{CNN}$ is first 3D-deconvolved into a  $128^3$ volume of $32$-dimensional features.
Each target view ray point $\x$ is then described with a positional embedding $\gamma(\x)$, and a latent feature $\z_\text{g}(\x) \in \mathbb{R}^{32}$ trilinearly sampled at the voxel grid location $\x$.
The rest is the same as in \textbf{MLP}.

Finally, the \textbf{Mesh} baseline uses the soft-rasterization of \cite{chen2019learning} as implemented in PyTorch3D~\cite{ravi2020pytorch3d} with the top-k face accumulation.
The scene encoding $\z_\text{CNN}$ is converted with a pair of linear layers to: 
(1) a set $\{v_i(\z) \in \mathbb{R}^3\}_{i=1}^{N_\text{vertex}}$ of 3D vertex locations of the object mesh, and 
(2) a $128\times 128$ UV map of the texture mapped to the surface of the mesh, which is rendered in order to evaluate the reconstruction losses from \cref{s:optimization}.
The mesh is initialized with an icosahedral sphere with 642 vertices.

\mysubsection{Quantitative Results}{quantitativeresults} 

\Cref{t:freicars} presents quantitative results on Freiburg Cars and the AMT Objects, respectively.
In terms of all perceptual metrics (\textbf{$\ell_1^{\text{RGB}}$}, \textbf{$\ell_1^{\text{VGG}}$}) as well as depth and IoU, our method is on par with the MLP on the \emph{train-test} split.
On the \emph{test} split, we outperform all other baselines in \textbf{$\ell_1^{\text{RGB}}$}, \textbf{$\ell_1^{\text{VGG}}$} and IoU on all 7 classes of AMT Objects and Freiburg Cars.
This indicates significantly better ability of our warp-conditioned embedding to generalize to previously unseen object instances.

We further find that our method is better at leveraging multiple source views $N_{\text{src}}>1$, outperforming all baselines for the $\ell_1^{\text{RGB}}$ error, see \Cref{t:viewablation}. When increasing the number of source images our method performance for all metrics improves whereas for all baselines it stays more or less constant. This further shows the effectiveness of the warp-conditioned embedding (WCR).

Regarding depth reconstruction ($\ell_1^{\text{Depth}}$), our method outperforms all alternatives on all datasets except the test split of Freiburg Cars, where we are 2nd after Mesh.
Here, we note that $\ell_1^{\text{Depth}}$ is only an approximate measure because: 1) the predicted depth is compared to the COLMAP-MVS estimate of depth~\cite{schoenberger2016mvs}, which tends to be noisy and; 
2) the scale ambiguity in SfM reconstructions that supervise learning leads to a significantly unconstrained problem of estimating the scale of a testing scene given a small number of source views, which is challenging to resolve for any method.

\mysubsection{Qualitative Results}{QualitativeResults} 
\refFig{SingleViewResults} provide qualitative comparisons for monocular novel-view synthesis.
It shows that our method produces significantly more detailed novel views, probably due to its ability to retrieve spatial encodings from the given source view.
\refFig{MultiViewResult} further demonstrates the reconstruction improvement when multiple source views $N^\text{src}>1$ are available.

%% file: table-results.tex
\begin{table*}[ht]
\small
\centering
\setlength{\tabcolsep}{4.8pt}
\begin{tabular}{lrrrrrrrrrrrrrrrr}
  &
  \multicolumn{8}{c}{\textbf{AMT}} &
  \multicolumn{8}{c}{\textbf{Freiburg Cars}} \\
  \cmidrule(lr){2-9} 
  \cmidrule(lr){10-17}
  &
  \multicolumn{4}{c}{\textbf{Train-test}} &
  \multicolumn{4}{c}{\textbf{Test}} & 
  \multicolumn{4}{c}{\textbf{Train-test}} &
  \multicolumn{4}{c}{\textbf{Test}} \\
  \cmidrule(lr){2-5} 
  \cmidrule(lr){6-9}
  \cmidrule(lr){10-13} 
  \cmidrule(lr){14-17}
  \textbf{Method} &
  \multicolumn{1}{c}{\textbf{$\ell_1^{\text{RGB}}$}} &
  \multicolumn{1}{c}{\textbf{\textbf{$\ell_1^{\text{VGG}}$}}} &
  \multicolumn{1}{c}{\textbf{IoU}} &
  \multicolumn{1}{c}{\textbf{\textbf{$\ell_1^{\text{Depth}}$}}} &
  \multicolumn{1}{c}{\textbf{$\ell_1^{\text{RGB}}$}} &
  \multicolumn{1}{c}{\textbf{\textbf{$\ell_1^{\text{VGG}}$}}} &
  \multicolumn{1}{c}{\textbf{IoU}} &
  \multicolumn{1}{c}{\textbf{\textbf{$\ell_1^{\text{Depth}}$}}} &
  \multicolumn{1}{c}{\textbf{$\ell_1^{\text{RGB}}$}} &
  \multicolumn{1}{c}{\textbf{\textbf{$\ell_1^{\text{VGG}}$}}} &
  \multicolumn{1}{c}{\textbf{IoU}} &
  \multicolumn{1}{c}{\textbf{\textbf{$\ell_1^{\text{Depth}}$}}} &
  \multicolumn{1}{c}{\textbf{$\ell_1^{\text{RGB}}$}} &
  \multicolumn{1}{c}{\textbf{\textbf{$\ell_1^{\text{VGG}}$}}} &
  \multicolumn{1}{c}{\textbf{IoU}} &
  \multicolumn{1}{c}{\textbf{\textbf{$\ell_1^{\text{Depth}}$}}} 
  \\
  \midrule
Mesh & 0.10 & 1.17 & 0.60 & 5.13 & 0.10 & 1.16 & 0.60 & 5.09 & 0.14 & 2.03 & 0.60 & 1.19 & 0.17 & 2.17 & 0.56 & \textbf{1.06} \\
Voxel & 0.06 & 1.05 & 0.78 & 2.14 & 0.09 & 1.13 & 0.66 & 3.07 & 0.05 & 1.58 & 0.89 & 0.59 & 0.16 & 2.05 & 0.51 & 2.18 \\
Voxel+MLP & 0.06 & 1.04 & 0.78 & 1.95 & 0.09 & 1.13 & 0.65 & 2.87 & 0.05 & 1.47 & 0.88 & \textbf{0.48} & 0.16 & 2.06 & 0.54 & 1.97 \\
MLP & 0.04 & 0.90 & 0.87 & 1.38 & 0.09 & 1.13 & 0.65 & 3.59 & \textbf{0.04} & \textbf{1.39} & 0.87 & 0.59 & 0.15 & 2.03 & 0.47 & 2.52 \\
\textbf{Ours} & \textbf{0.03} & \textbf{0.86} & \textbf{0.88} & \textbf{1.31} & \textbf{0.05} & \textbf{0.93} & \textbf{0.83} & \textbf{1.90} &
\textbf{0.04} & \textbf{1.39} & \textbf{0.90} & \textbf{0.48} & \textbf{0.12} & \textbf{1.89} & \textbf{0.62} & 1.60 \\
\bottomrule
\end{tabular}%
\vspace{-0.7em}%
\caption{%
\textbf{Novel-view synthesis on AMT Objects and Freiburg Cars}. 
Each row evaluates either a baseline or \textbf{our} method.
Results are reported for two perceptual metrics $\ell_1^{\text{RGB}}$, $\ell_1^{\text{VGG}}$, depth error  $\ell_1^{\text{Depth}}$, and intersection-over-union (IoU). For training we randomly selected between 1 and 7 source images. For testing we separately calculated the error metrics for 1, 3, 5 and 7 source images respectively and provide the average among those. For a more detailed evaluation we refer to the supplemental. Lower is better for \textbf{$\ell_1^{\text{RGB}}$, $\ell_1^{\text{VGG}}$}, and \textbf{$\ell_1^{\text{Depth}}$}, whereas higher is better for \textbf{IoU}.
The best result is \textbf{bolded}.\label{t:freicars}%
}\vspace{-0.5em}%
\end{table*}

%% file: table-viewablation.tex
\begin{table*}[ht]
\small
\centering
\setlength{\tabcolsep}{4.8pt}
\begin{tabular}{llrrrrrrrrrrrrrrrr}
  & &
  \multicolumn{8}{c}{\textbf{AMT}} &
  \multicolumn{8}{c}{\textbf{Freiburg Cars}} \\
  \cmidrule(lr){3-10} 
  \cmidrule(lr){11-18}
  & &
  \multicolumn{4}{c}{\textbf{Train-test}} &
  \multicolumn{4}{c}{\textbf{Test}} & 
  \multicolumn{4}{c}{\textbf{Train-test}} &
  \multicolumn{4}{c}{\textbf{Test}} \\
  \cmidrule(lr){3-6} 
  \cmidrule(lr){7-10}
  \cmidrule(lr){11-14} 
  \cmidrule(lr){15-18}
  & \textbf{Method} &
  \multicolumn{1}{c}{1} &
  \multicolumn{1}{c}{3} &
  \multicolumn{1}{c}{5} &
  \multicolumn{1}{c}{7} &
  \multicolumn{1}{c}{1} &
  \multicolumn{1}{c}{3} &
  \multicolumn{1}{c}{5} &
  \multicolumn{1}{c}{7} &
  \multicolumn{1}{c}{1} &
  \multicolumn{1}{c}{3} &
  \multicolumn{1}{c}{5} &
  \multicolumn{1}{c}{7} &
  \multicolumn{1}{c}{1} &
  \multicolumn{1}{c}{3} &
  \multicolumn{1}{c}{5} &
  \multicolumn{1}{c}{7}
  \\
  \midrule
 \multirow{5}{*}{\rotatebox{90}{\textbf{$\ell_1^{\text{RGB}}$}}} & 
Mesh & .096 & .096 & .096 & .096 & .102 & .102 & .102 & .102 & .141 & .141 & .140 & .140 & .166 & .166 & .166 & .166 \\
& Voxel & .062 & .061 & .061 & .061 & .091 & .091 & .091 & .091 & .055 & .055 & .055 & .054 & .159 & .159 & .158 & .158 \\
& Voxel+MLP & .059 & .059 & .058 & .059 & .090 & .090 & .090 & .090 & .045 & .045 & .045 & .045 & .158 & .157 & .158 & .157 \\
& MLP & \textbf{.037} & .036 & .036 & .036 & .088 & .088 & .088 & .088 & \textbf{.041} & \textbf{.041} & \textbf{.041} & .041 & .152 & .152 & .152 & .152 \\
& \textbf{Ours} & .038 & \textbf{.032} & \textbf{.031} & \textbf{.030} & \textbf{.058} & \textbf{.046} & \textbf{.043} & \textbf{.042} & .046 & \textbf{.041} & \textbf{.041} & \textbf{.040} & \textbf{.130} & \textbf{.120} & \textbf{.115} & \textbf{.114} \\
  
\bottomrule
\end{tabular}%
\vspace{-0.7em}%
\caption{\label{t:viewablation}
We evaluate the impact of increasing the number of source views during test time for the $\ell_1^{\text{RGB}}$ metric. 
Target renders and the corresponding metrics are produced for 1, 3, 5 and 7 source images.
The best result is \textbf{bolded} where lower is better.
}\vspace{-0.5em}%
\end{table*}

%% file: conclusions.tex
\mysection{Discussion and conclusions}{Discussion}

\paragraph{Limitations.}

Even though our method outperforms baselines on the vast majority of metrics and datasets, there are still several limitations. 
First, the execution of the deep MLP at every 3D ray-location in a rendered frame is relatively slow (depending on the number of source views rendering takes between 3 and 8 sec for a $128 \times 256$ image on average), which makes a real-time deployment challenging. 
Secondly, due to our template-free approach, the object silhouettes can be blurry. 
Lastly, despite no manual labeling is necessary, our 
method still relies on segmentation masks that were automatically generated with Mask-RCNN.

\paragraph{Conclusions.}

In this paper, we have presented a method that is able to reconstruct category-specific 3D shape and appearance from videos of object categories in the wild alone, without requiring manual annotations.
We demonstrated that our main contribution, Warp-Conditioned Ray Embedding, can successfully deal with the inherent ambiguities present in the video SfM reconstructions that provide our supervisory signal, outperforming alternatives on a novel dataset of crowd-sourced object videos.
Future work could include decomposition of shape, appearance and lighting allowing for more control over the rendered images.

%% file: supplemental.tex
\begin{strip}%
 \centering
 \Large
 \textbf{Unsupervised Learning of 3D Object Categories from Videos in the Wild\\
 \vspace{0.3cm} \textit{Supplementary material}
 }
\vspace{0.5cm}
\end{strip}

\input{table-viewablation-suppl}

\mysection{Additional implementation details}{WCEDetails}

In this section, we provide more detailed information about the dense image descriptors $\Phi$ as well as the neural radiance field $\Psi$. Furthermore, we give more insights into the training process.   

\mysubsection{Dense image descriptors}{DenseDescriptors}
\label{s:dense-decsriptor}
This section describes in more detail the dense pixel-wise embeddings $\Phi(I_t)$ introduced in Section 3.3 in the main paper.

For a given source image $I_t$, the embedding 
field $\Phi(I_t)$ 
is composed of 3 different types of features:
1) learned $5 \cdot 32$-dimensional dense pixel-wise features output by a deep convolutional encoder network $\Phi_\text{U-Net}$, 
2) raw image rgb colors $I_t \in \mathbb{R}^{3 \times H \times W}$,
and 3) the segmentation mask $m_t \in \mathbb{R}^{1 \times H \times W}$.

\paragraph{Dense feature extractor $\Phi_\text{U-Net}$.}
The architecture of the U-Net inside $\Phi_\text{U-Net}$ is defined as follows (a detailed visualisation is present in \refFig{Encoder}).
A source image $ I^{\text{src}} \in \mathbb{R}^{3 \times H \times W}$, masked by $ m^{\text{src}} $ (retrieved from Mask-RCNN), 
is fed into a ResNet-50 which returns spatial features from intermediate convolutional layers (\emph{layer1, layer2, layer3, layer4, layer5}), and the final linear ResNet layer which outputs global features $\z_\text{CNN}$, i.e. non-spatial. 
Each feature layer including the global one is then passed through a 1x1 convolution to equalize the size of all feature channels to 32.
The spatial features are further bilinearly upsampled to the spatial size of the source image and concatenated along the channel dimension to create a dense embedding field 
$\Phi_\text{U-Net}(I_t) \in \mathbb{R}^{5\cdot32 \times H \times W}$.

\paragraph{Neural radiance field $\Psi$.}

\myfigure{Decoder}{The neural radiance field $\Psi$ is represented by an MLP. It takes as input the warp-conditioned embedding $\z(\x)$, the harmonic positional embedding $\gamma(\x)$ and to account for view point variations the harmonic directional embedding $\gamma(\br)$. It returns the rgb and opacity values.}

Our scene is represented by a neural radiance field $\Psi$ similar to \cite{mildenhall2020nerf} with the only difference that we additionally condition the field with our warp-conditioned ray embedding, see \refFig{Decoder}.

\mysubsection{Training details}{TrainingDetails}

We trained both the U-Net encoder $\Phi_\text{U-Net}$ and the neural radiance field $\Psi$ with Adam optimizer. We set the batch size to 8 and the learning rate to 1e-4. Our method as well as all baselines were trained on an NVIDIA Tesla V100 for 7 days.
For all raymarching baselines and our method, we shoot 1024 rays per iteration through random image pixels in Monte-Carlo fashion. For each ray we first uniformly sample 128 times along the ray in order to retrieve a coarse rendering (voxel or mlp based depending on the method used). In the second pass we sample each ray 128 times based on probabilistic importance sampling following \cite{mildenhall2020nerf}. 

For the mesh baseline we shoot rays for each pixel per iteration and use soft rasterization to predict the surface intersection. In addition to the losses used for the other baselines as well as our method, we additionally use a negative IoU loss $L_{iou}$, a Laplacian loss $L_{lap}$ and smoothness loss $L_{sm}$ according to \cite{ravi2020pytorch3d} and weighted them with 1.0, 19.0, 1.0 respectively.

\mycfigure{Encoder}{
The input to the dense feature extractor $\Phi$ is a source image from a given view. It first makes use of a ResNet-50 ($\Phi_\text{U-Net}$) to retrieve the layer-wise features.
Then, each layer is independently fed to a 1x1 convolution followed by bilinear upsmapling to the original input resolution. The resulting feature blocks are concatenated with the input image $I^{src}$ and its corresponding object mask $m^{src}$. In case there are multiple source images available, this process is repeated for each of them. Once all per-view features are obtained the warp-conditioned ray embedding is retrieved after applying the view-aggregation.
}

\mysection{Additional qualitative results}{AddResults}

Additional qualitative results are available presented in \refFig{MultiViewResultsSuppl} and \refFig{SingleViewResultsSuppl}. Also, we provide more qualitative results
on our project webpage: \url{https://henzler.github.io/publication/unsupervised_videos/}.
The page contains comparison of our method to baselines by showing the scenes from the train-test or test subsets rendered from a viewpoint that rotates around the object of interest.

\mysection{Test-time view ablation}{ViewAblation}

Furthermore, we also provide a view ablation of our method at test time. Recall that we randomly sample between 1 and 7 source images during training. During test time we evaluated our method separately on 1, 3, 5 and 7 views as input. In the main paper we provide an average of those numbers. In \Cref{t:viewablation-suppl} we give insight into how changing the number of source views affects performance. Not surprisingly, increasing the numbers of source views consistently improves all metrics.

\mycfigure{SingleViewResultsSuppl}{In reach row, a single source image (1st column) is processed by one of the evaluated methods (Mesh, Voxel, MLP+Voxel, MLP, \textbf{Ours} - columns 2 to 6) to generate a prescribed target view (last column).
We show results on the test split.
}

\mycfigure{MultiViewResultsSuppl}{\textbf{Reconstruction with multiple source views.}
The top row for each object shows all available source images (columns 1-7) for a given target image (top right).
The bottom row contains results conditioned on 1, 3, 5 or 7 source images. In addition to the rendered new RGB views we also provide shaded surface renderings.}

%% file: table-viewablation-suppl.tex
\begin{table*}[ht]
\small
\centering
\setlength{\tabcolsep}{3.0pt}
\begin{tabular}{llrrrrrrrrrrrrrrrr}
  & &
  \multicolumn{8}{c}{\textbf{AMT}} &
  \multicolumn{8}{c}{\textbf{Freiburg Cars}} \\
  \cmidrule(lr){3-10} 
  \cmidrule(lr){11-18}
  & &
  \multicolumn{4}{c}{\textbf{Train-test}} &
  \multicolumn{4}{c}{\textbf{Test}} & 
  \multicolumn{4}{c}{\textbf{Train-test}} &
  \multicolumn{4}{c}{\textbf{Test}} \\
  \cmidrule(lr){3-6} 
  \cmidrule(lr){7-10}
  \cmidrule(lr){11-14} 
  \cmidrule(lr){15-18}
  & \textbf{Method} &
  \multicolumn{1}{c}{1} &
  \multicolumn{1}{c}{3} &
  \multicolumn{1}{c}{5} &
  \multicolumn{1}{c}{7} &
  \multicolumn{1}{c}{1} &
  \multicolumn{1}{c}{3} &
  \multicolumn{1}{c}{5} &
  \multicolumn{1}{c}{7} &
  \multicolumn{1}{c}{1} &
  \multicolumn{1}{c}{3} &
  \multicolumn{1}{c}{5} &
  \multicolumn{1}{c}{7} &
  \multicolumn{1}{c}{1} &
  \multicolumn{1}{c}{3} &
  \multicolumn{1}{c}{5} &
  \multicolumn{1}{c}{7}
  \\
  \midrule
  \multirow{5}{*}{\rotatebox{90}{\textbf{$\ell_1^{\text{VGG}}$}}} & 
Mesh & 1.163 & 1.167 & 1.168 & 1.169 & 1.160 & 1.161 & 1.163 & 1.163 & 2.030 & 2.029 & 2.028 & 2.023 & 2.170 & 2.168 & 2.166 & 2.167 \\
& Voxel & 1.052 & 1.051 & 1.051 & 1.051 & 1.127 & 1.127 & 1.127 & 1.127 & 1.581 & 1.581 & 1.580 & 1.580 & 2.050 & 2.050 & 2.046 & 2.046 \\
& Voxel+MLP & 1.041 & 1.040 & 1.040 & 1.040 & 1.131 & 1.130 & 1.130 & 1.130 & 1.469 & 1.468 & 1.468 & 1.468 & 2.067 & 2.063 & 2.063 & 2.064 \\
& MLP & \textbf{0.900} & 0.899 & 0.899 & 0.899 & 1.130 & 1.130 & 1.130 & 1.131 & \textbf{1.391} & 1.389 & 1.389 & 1.389 & 2.027 & 2.025 & 2.024 & 2.025 \\
& \textbf{Ours} & 0.905 & \textbf{0.846} & \textbf{0.837} & \textbf{0.832} & \textbf{1.007} & \textbf{0.921} & \textbf{0.896} & \textbf{0.883} & 1.450 & \textbf{1.381} & \textbf{1.372} & \textbf{1.359} & \textbf{1.945} & \textbf{1.897} & \textbf{1.874} & \textbf{1.863} \\
 \midrule
 \multirow{5}{*}{\rotatebox{90}{\textbf{IoU}}} & 
Mesh & 0.599 & 0.599 & 0.599 & 0.598 & 0.598 & 0.598 & 0.598 & 0.598 & 0.601 & 0.604 & 0.605 & 0.606 & 0.556 & 0.556 & 0.556 & 0.556 \\
& Voxel & 0.776 & 0.777 & 0.777 & 0.777 & 0.660 & 0.660 & 0.660 & 0.661 & 0.891 & 0.892 & 0.892 & 0.893 & 0.517 & 0.511 & 0.509 & 0.510 \\
& Voxel+MLP & 0.775 & 0.776 & 0.777 & 0.776 & 0.652 & 0.654 & 0.654 & 0.654 & 0.878 & 0.878 & 0.878 & 0.878 & 0.540 & 0.541 & 0.542 & 0.541 \\
& MLP & \textbf{0.871} & 0.871 & 0.872 & 0.872 & 0.654 & 0.653 & 0.653 & 0.653 & 0.872 & 0.872 & 0.872 & 0.872 & 0.472 & 0.470 & 0.472 & 0.471 \\
& \textbf{Ours} & 0.866 & \textbf{0.884} & \textbf{0.886} & \textbf{0.889} & \textbf{0.774} & \textbf{0.788} & \textbf{0.787} & \textbf{0.787} & \textbf{0.889} & \textbf{0.897} & \textbf{0.898} & \textbf{0.897} & \textbf{0.600} & \textbf{0.624} & \textbf{0.629} & \textbf{0.632} \\
\midrule
\multirow{5}{*}{\rotatebox{90}{\textbf{$\ell_1^{\text{Depth}}$}}} & 
Mesh & 5.138 & 5.119 & 5.128 & 5.130 & 5.100 & 5.101 & 5.090 & 5.086 & 1.202 & 1.185 & 1.178 & 1.177 & \textbf{1.062} & \textbf{1.061} & \textbf{1.063} & \textbf{1.063} \\
& Voxel & 2.150 & 2.141 & 2.140 & 2.141 & 3.069 & 3.064 & 3.067 & 3.065 & 0.591 & 0.590 & 0.585 & 0.583 & 2.133 & 2.181 & 2.207 & 2.200 \\
& Voxel+MLP & 1.958 & 1.942 & 1.942 & 1.941 & 2.881 & 2.868 & 2.861 & 2.864 & \textbf{0.478} & 0.479 & 0.479 & 0.479 & 1.972 & 1.979 & 1.968 & 1.968 \\
& MLP & \textbf{1.389} & 1.378 & 1.377 & 1.377 & 3.583 & 3.587 & 3.590 & 3.593 & 0.595 & 0.593 & 0.594 & 0.593 & 2.521 & 2.530 & 2.519 & 2.520 \\
& \textbf{Ours} & 1.593 & \textbf{1.291} & \textbf{1.201} & \textbf{1.172} & \textbf{2.186} & \textbf{1.847} & \textbf{1.802} & \textbf{1.776} & 0.535 & \textbf{0.467} & \textbf{0.457} & \textbf{0.453} & 1.606 & 1.595 & 1.589 & 1.603 \\
  
\bottomrule
\end{tabular}%
\vspace{-0.7em}%
\caption{\label{t:viewablation-suppl}
We complement the evaluation of the impact of the number of source views during test time for the metrics: \textbf{$\ell_1^{\text{VGG}}$}, \textbf{$\ell_1^{\text{Depth}}$}, \textbf{IoU}. We report results for 1, 3, 5 and 7 source images.
The best result is \textbf{bolded} where lower is better for \textbf{$\ell_1^{\text{VGG}}$}, \textbf{$\ell_1^{\text{Depth}}$} and higher is better for \textbf{IoU}.
}\vspace{-0.5em}%
\end{table*}